\documentclass[letterpaper, 10 pt, conference]{ieeeconf}  

\IEEEoverridecommandlockouts                              

\usepackage{xcolor}  
\definecolor{cvprblue}{rgb}{0.21,0.49,0.74}
\definecolor{lightblue}{RGB}{230, 242, 255}

\usepackage[pagebackref,breaklinks,colorlinks,allcolors=cvprblue]{hyperref}
\usepackage{graphicx}
\usepackage{wrapfig}
\usepackage{multirow}
\usepackage{booktabs}
\usepackage[table]{xcolor}
\usepackage{relsize}
\usepackage[most]{tcolorbox}
\usepackage{cuted}
\usepackage{caption}
\usepackage{amsmath} 
\usepackage{amssymb} 
\usepackage{framed}
\definecolor{mygreen}{RGB}{0,150,0}
\definecolor{myred}{RGB}{200,0,0}
\usepackage{pifont}
\usepackage{url}
\usepackage[misc]{ifsym}

\title{\LARGE \bf
HiMemVLN: Enhancing Reliability of Open-Source Zero-Shot Vision-and-Language Navigation with \underline{Hi}erarchical \underline{Mem}ory System
}

\author{
Kailin Lyu$^{1,9*}$, Kangyi Wu$^{2}$, Pengna Li$^{2}$, Xiuyu Hu$^{3}$,
Qingyi Si$^{4}$, Cui Miao$^{5}$, Ning Yang$^{1,10}$, Zihang Wang$^{6}$,\\
Long Xiao$^{1}$, Lianyu Hu$^{7,(\textrm{\Letter})}$, Jingyuan Sun$^{8,(\textrm{\Letter})}$, Ce Hao$^{9}$%
\thanks{*The work was conducted during the internship of Kailin Lyu  (lvkailin2024@ia.ac.cn) at Huawei Technologies Co., Ltd., Beijing, China.}
\thanks{$^{1}$Kailin Lyu is with the Institute of Automation, Chinese Academy of Sciences, and the Zhongguancun Academy, Beijing, China}%
\thanks{$^{2}$Kangyi Wu and Pengna Li are with the Institute of Artificial Intelligence and Robotics, Xi'an Jiaotong University, Xi'an, China}%
\thanks{$^{3}$Xiuyu Hu is with the School of Transportation, Tongji University, Shanghai, China}%
\thanks{$^{4}$Qingyi Si is with JD.com, Beijing, China}%
\thanks{$^{5}$Cui Miao is with the Institute of National University of Defense Technology, Changsha, China}%
\thanks{$^{6}$Zihang Wang is with Southeast University, Nanjing, China}%
\thanks{$^{7}$Lianyu Hu is with Nanyang Technological University, Singapore}%
\thanks{$^{8}$Jingyuan Sun is with Huawei Technologies Co., Ltd., Shanghai, China}%
\thanks{$^{9}$Ce Hao is with the Zhongguancun Academy, Beijing, China}%
\thanks{$^{10}$Ning Yang is with the School of Intelligent Science and Technology, Nanjing University, Suzhou, China, and the Institute of Automation, Chinese Academy of Sciences, Beijing, China}%
\thanks{\textrm{\Letter} Lianyu Hu (huhly2021@tju.edu.cn) and Jingyuan Sun (sunjingyuan1@huawei.com) are the corresponding authors.}
}

\begin{document}

\maketitle
\thispagestyle{empty}
\pagestyle{empty}

\begin{abstract}
LLM-based agents have demonstrated impressive zero-shot performance in vision-language navigation (VLN) task. However, most zero-shot methods primarily rely on closed-source LLMs as navigators, which face challenges related to high token costs and potential data leakage risks. Recent efforts have attempted to address this by using open-source LLMs combined with a spatiotemporal CoT framework, but they still fall far short compared to closed-source models. In this work, we identify a critical issue, Navigation Amnesia, through a detailed analysis of the navigation process. This issue leads to navigation failures and amplifies the gap between open-source and closed-source methods. To address this, we propose HiMemVLN, which incorporates a Hierarchical Memory System into a multimodal large model to enhance visual perception recall and long-term localization, mitigating the amnesia issue and improving the agent’s navigation performance. Extensive experiments in both simulated and real-world environments demonstrate that HiMemVLN achieves nearly twice the performance of the open-source state-of-the-art method. The code is available at \url{https://github.com/lvkailin0118/HiMemVLN}.
\end{abstract}

\section{Introduction}
\label{intro}
Vision-and-Language Navigation (VLN) is a fundamental task in Embodied AI \cite{vln1}. It requires the agent to navigate in novel environments according to natural language instructions. Although supervised learning has achieved strong results on VLN benchmark datasets, its dependence on large-scale human annotations substantially limits generalization to unseen environments \cite{etpnav,cma,bevbert}. With the rapid progress of large language
models (LLMs) \cite{llm2}, zero-shot navigation driven by commonsense reasoning has become a promising research direction. However, many state-of-the-art approaches rely on closed-source cloud-hosted models \cite{canav,smartway}, such as GPT-4o \cite{gpt4o}. Despite their strong reasoning capability, cloud dependence introduces practical barriers in real deployments. As illustrated in Fig.~\ref{intro}(a), these barriers include high API costs and unpredictable communication latency. More critically, uploading real-time visual streams that may contain sensitive spatial information to external servers raises privacy and security risks, which hinders deployment in home service and industrial robotics. Consequently, deploying open-source LLMs (e.g., Llama \cite{llama} and Qwen \cite{qwen2}) directly on the robot has become an important pathway toward secure and autonomous embodied agents.

\begin{figure*}[!t]
\centering
\includegraphics[width=\textwidth]{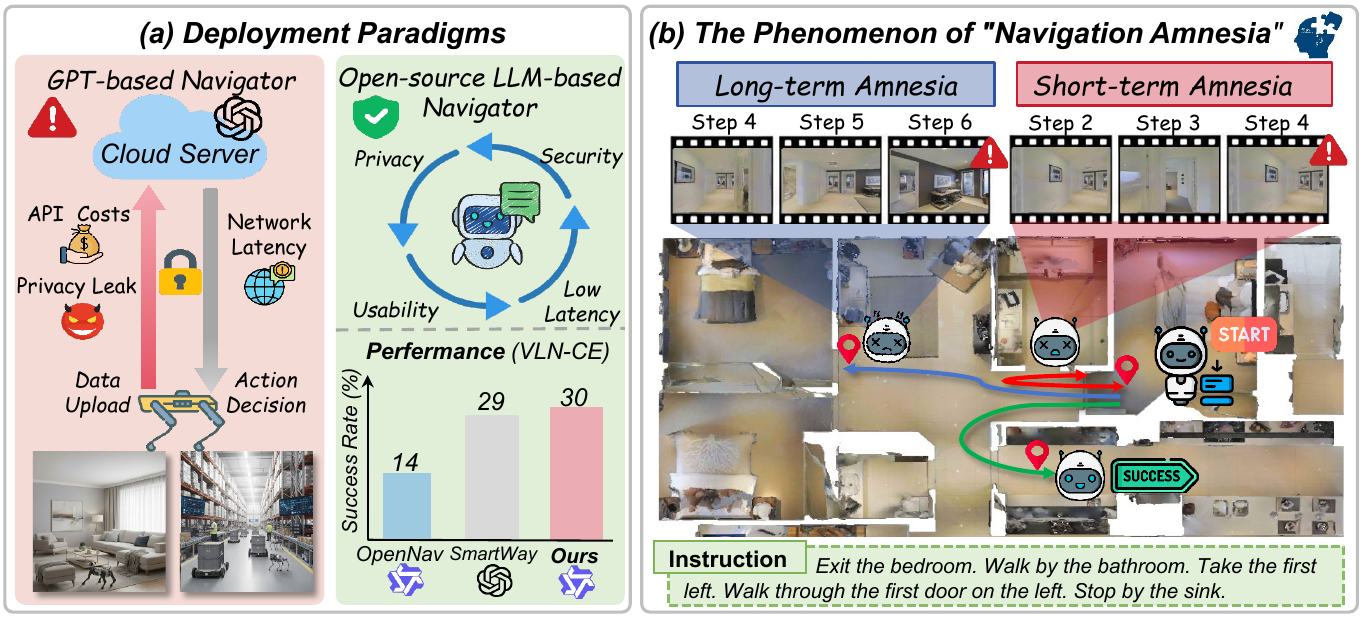}
\caption{(a) Comparison between GPT-based Navigator and open-source LLM-based Navigator. (b) The Phenomenon of "Navigation Amnesia". Short term amnesia leads the agent to enter local loops or perform redundant exploration, whereas long term amnesia causes it to forget global instruction constraints, making decisions drift from the intended route.}
\label{intro}
\end{figure*}

Although prior work has explored using open-source LLM-based Navigator in continuous environments, a substantial performance gap remains relative to closed-source models \cite{opennav}. We argue that a primary driver of this gap is insufficient memory awareness of environmental state and task progress, which leads agents to exhibit an “navigation amnesia” phenomenon during multi step. As shown in Fig.~\ref{intro}(b), from the perspective of short term perception, existing approaches often convert continuous, dynamic visual streams into discrete textual descriptions, which yields incomplete perception and impoverished scene understanding \cite{opennav,navgpt,discussnav}. In continuous environments, adjacent viewpoints or visually similar scenes frequently produce near duplicate descriptions that fail to preserve discriminative visual cues. As a result, the agent struggles to reliably identify the current state, and may enter local loops or trigger redundant exploration. From the perspective of long horizon planning, open-source models typically have smaller context windows and weaker implicit memory capacity than closed-source models \cite{openclose1,openclose2}, making it difficult to maintain coherent state estimates over extended trajectories. As navigation steps increases and reasoning noise accumulates, the agent is prone to forgetting global constraints in the instruction, causing decisions to drift from the intended route and reducing the ability to self correct.

To overcome this problem, we resort to neurobiology \cite{theory2}, specifically the Hippocampal Memory Indexing Theory \cite{theory1}:
\begin{tcolorbox}[colframe=black!50, colback=cvprblue!8, boxrule=1.5pt, arc=2mm, top=4pt, bottom=4pt, left=4pt, right=4pt,  boxsep=1pt]
\raisebox{-0.2\baselineskip}{\includegraphics[height=1.2\baselineskip]{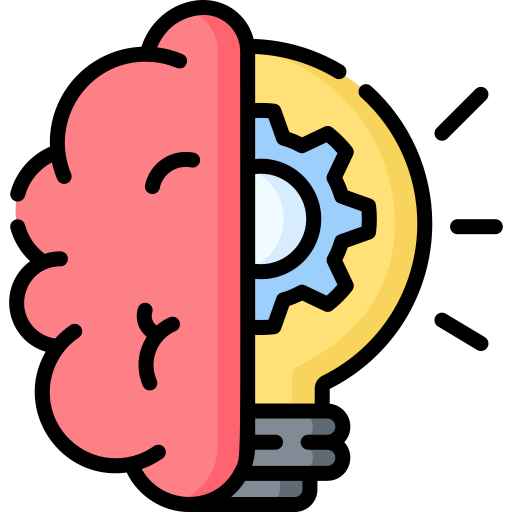}} \textit{Distinct neural circuits coordinate short term and long term memory to support spatial cognition and decision making. Short term memory relies on rapid perceptual processes that encode current visual input, whereas long term memory consolidates and retrieves abstract concepts to maintain a coherent global representation.}
\end{tcolorbox} 
\noindent While this cognitive theory reveals the essence of human cognition, it can be smoothly translated into an architectural principle of VLN: short term memory is visually dominant, enhancing the agent’s immediate perception and state representation of the current scene, while long term memory is semantically dominant, providing stable semantic grounding, generalized knowledge, and contextual information to complete the perception to reasoning to decision pipeline.

Based on such inspiration, we propose \textbf{HiMemVLN}, a cognitively grounded framework that hierarchically integrates short term and long term memory into VLN. To address short term memory deficiencies, we introduce a \textit{\textbf{Short Term Localer System}}. Unlike prior approaches that rely solely on textual histories, we design a Visual Graph Memory module that stores visual features from previously visited viewpoints and aligns the current observation with historical nodes at each navigation step to explicitly localize the agent state. This mechanism enables visual retrospection, allowing the agent to reliably detect revisits and to impose soft constraints that suppress redundant exploration. To mitigate long horizon planning drift, we further propose a \textit{\textbf{Long Term Globaler System}}. At the beginning of an episode, this system extracts a global navigation schema from the instruction, including the principal direction and the target landmark. During navigation, the Globaler continuously recalls the trajectory history and calibrates current decisions so that local actions remain consistent with the global goal, thereby improving long range reasoning consistency. We perform comprehensive experiments to assess the effectiveness of HiMemVLN in both simulated and real-world environments, validating its performance across diverse settings. In this work, our main contributions are as follows:

\begin{itemize}
  \item We propose \textbf{HiMemVLN}, a hierarchical memory based VLN framework for open source LLMs that improves zero shot navigation reliability under privacy preserving and deployment oriented constraints.
  \item Inspired by cognitive theory, HiMemVLN combines a short term visually dominant Localer for state localization with a long term semantically dominant Globaler for persistent goal grounding, improving long horizon consistency and decision stability.
  \item Across simulated and real environments, HiMemVLN consistently outperforms Open-Nav over diverse scenes, with the best success rate approaching a two fold improvement and a smaller gap to closed-source models.
\end{itemize}

\section{Related Work}

\subsection{Vision-and-Language Navigation}
\label{Vision-and-Language Navigation}

Vision-and-Language Navigation (VLN) is a representative task in the field of embodied AI, requiring an agent to follow instructions grounded in visual observations to reach a target location \cite{vln1}. Early research primarily adopted a discrete VLN setting based on the MP3D simulator\cite{mp3d}, in which the agent navigates within predefined graph structures. To bridge the gap between simulated environments and real world applications, Krantz et al \cite{vlnce}. introduced a benchmark that extends the VLN task from discrete environments to continuous environments (CE), making it more representative of real world navigation scenarios.

Early VLN approaches employed LSTM models for cross modal alignment and navigation policy learning \cite{lstm1,lstm2}. Building upon these methods, subsequent research proposed a variety of improvements, including stronger representation learning techniques \cite{rep1,rep2}, progress monitoring mechanisms \cite{canav,navgpt}, and reinforcement learning based methods \cite{regnav}. Beyond architectural improvements, recent work has also focused on enhancing the agent’s generalization ability in novel environments through data augmentation. In contrast to these approaches, we aim to explore a training free solution that performs zero shot VLN in continuous environments by using open-source large language models as navigators.

\subsection{Navigation with Large Language Models}
\label{Navigation with Large Language Models}

In navigation tasks, there is often a wealth of visual and semantic information involved. Hence, many LLM-based navigation methods have emerged in recent years. For example, NavGPT \cite{navgpt} leverages GPT-4 to generate navigation decisions from textualized visual observations and trajectory history. InstructNav \cite{instructnav} employs GPT-4V with panoramic inputs to estimate heading values and decomposes navigation into subgoals for efficient execution. LLMPlanner \cite{llmplanner} produces multi-step plans that are dynamically refined using object detection signals and predefined rules.

Recent approaches integrate large language models as central reasoning modules to enable zero-shot VLN \cite{canav,smartway}. However, most methods rely on GPT-4, requiring frequent API calls, incurring high costs, and raising privacy concerns due to cloud-based scene transmission. Although OpenNav \cite{opennav} explores zero-shot navigation with open-source models, its performance remains substantially below that of closed-source counterparts due to architectural limitations. To overcome these issues, we propose a hierarchical memory-based VLN framework that leverages open-source LLMs as navigators and enhances environmental understanding and task planning capabilities through structured design, enabling more reliable open-source zero-shot navigation.

\section{Preliminaries}
\label{Preliminaries}

\subsection{Problem Definition}
\label{Problem Definition}
We address the zero-shot VLN-CE task, where an agent navigates within a continuous 3D space \( E \) based on natural language instructions \cite{vlnce}. At each position, the agent receives panoramic RGBD observations taken at evenly spaced viewpoints (e.g., 0°, 30°, ..., 330°), obtaining 12 RGB images and 12 depth images, denoted as \( I = \{ (I_i^{rgb}, I_i^{depth}) \mid i = 1, \ldots, 12 \} \), where \( I_i^{rgb} \in \mathbb{R}^{H \times W \times 3} \) and \( I_i^{depth} \in \mathbb{R}^{H \times W} \). Additionally, for each trial, the agent receives corresponding linguistic instructions, represented by the language embeddings \( L = \{ l_1, l_2, \ldots, l_n \} \), where each \( l_i \) denotes a token in the instruction. The instruction specifies how the agent should navigate from the starting position \( x_{start} \in E \) to the goal position \( x_{goal} \in E \). The agent can perform low-level actions, such as MOVE FORWARD (0.25 m), TURN LEFT/RIGHT (30°), and STOP. An episode is considered successful if the agent stops within a certain distance of the target location.

\subsection{Waypoint Prediction}
\label{Waypoint Prediction}

The waypoint predictor generates candidate waypoints based on the current observation, bridging the gap between discrete environments and continuous space. We employ a waypoint predictor similar to OpenNav \cite{opennav}, which uses a transformer-based model to predict waypoints. For any given position in an open environment, the agent captures RGB and depth data, each consisting of 12 single-view images taken at 30-degree intervals. RGB images are encoded using a ResNet-50 pre-trained on ImageNet \cite{resnet}, producing \( f^{rgb} \), while depth images are encoded using another ResNet-50 pre-trained for point-goal navigation \cite{depth}, resulting in \( f^{depth}_i \). These features are fused through a non-linear layer \( W_m \), resulting in a combined feature \( f^{rgbd}_i \). The fused feature vectors \( f^{rgbd}_i \) are processed by a two-layer transformer network to model spatial relationships between views. The network’s self-attention is restricted to adjacent views to enhance spatial reasoning:
\begin{equation}
\tilde{f}_{rgbd_i} = \text{Transformer}\left(\{f_{rgbd_{i-1}}, f_{rgbd_i}, f_{rgbd_{i+1}}\}\right),
\end{equation}
The output from the transformer is used to generate a heatmap of potential waypoints, which is then refined using non-maximum suppression (NMS) to extract the top \( K \) navigable waypoints, with \( K \) set to 5. Each waypoint includes the angle (in radians) and distance (in meters) from the agent to the waypoint, providing precise directional and spatial information necessary for navigation.

\section{Method}
\label{method}
\begin{figure}[!t] 
    \centering 
    \includegraphics[width=\columnwidth]{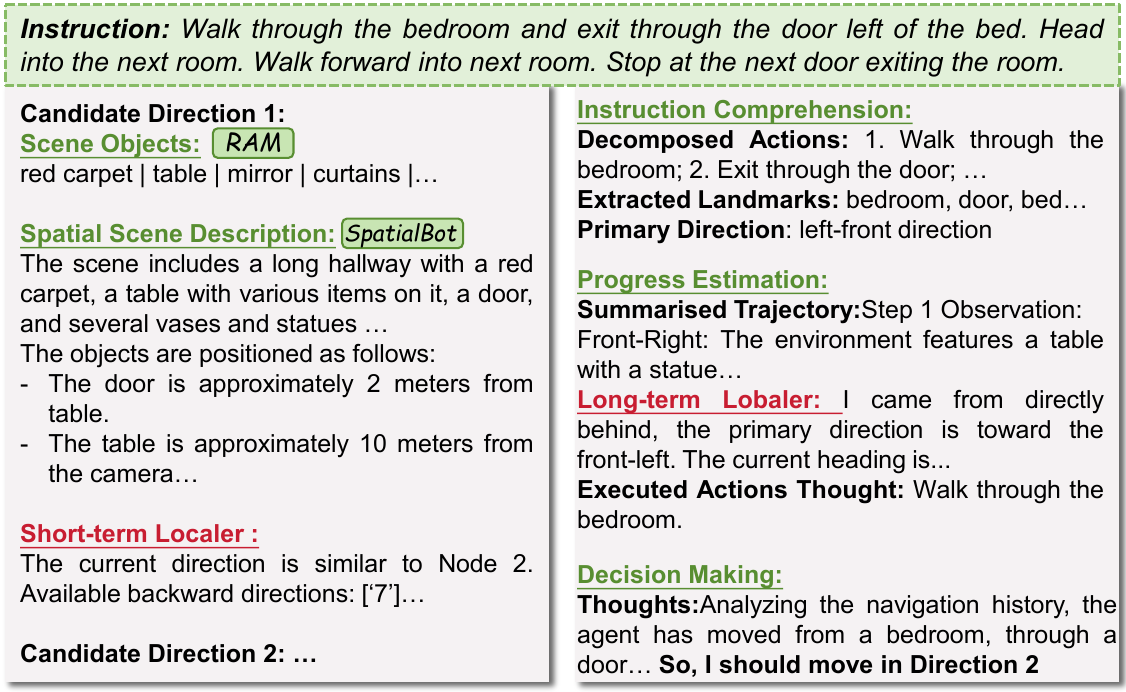}    
    \caption{\textbf{Overview of HiMemVLN.} Built upon a MLLM, HiMemVLN integrates short-term and long-term memory systems to accomplish task execution through a closed-loop memory-reasoning-execution process.}
    \vspace{-3mm}
    \label{Overview} 
\end{figure}

The overall pipeline of the proposed HiMemVLN framework is illustrated in Fig~\ref{Overview}. \ding{182} First, the agent perceives the current environment. At each step, it receives panoramic multi-view RGB and depth observations. A waypoint predictor estimates feasible rotations and translations to transform the continuous action space into a compact set of navigable candidates. For each candidate, RAM \cite{ram} and SpatialBot \cite{spatialbot} extract semantic objects and spatial structures, which are converted into structured textual descriptions. To address the short-term amnesia issue discussed in Sec.~\ref{intro}, we introduce the Short-Term Localer, which formalizes visual perception during navigation into a short-term memory representation. By maintaining an online visual graph memory, it localizes observations, detects revisits, and imposes soft exploration constraints, thereby stabilizing short-term state estimation and reducing local loops. \ding{183} Subsequently, the agent performs planning conditioned on the given instruction, textualized observations, and navigation history. An MLLM-based navigator conducts structured reasoning by decomposing subgoals, estimating progress, and selecting the next waypoint according to instruction alignment, spatial feasibility, and exploration priority. To address trajectory drift caused by long-term amnesia, we introduce the Long-Term Globaler to correct navigation at the level of global intent. It formalizes high-level semantic information, including navigational intent, into a persistent long-term memory representation, and performs reflective reasoning during inference to track the overall instruction and mitigate long-horizon drift. \ding{184} Finally, the selected high-level action is translated by a low-level planner into executable motion commands, producing new observations and thereby closing the memory---reasoning---execution loop. We provide a detailed description of the short-term and long-term systems as follows.

\begin{figure*}[!t] 
    \centering 
    \includegraphics[width=\textwidth]{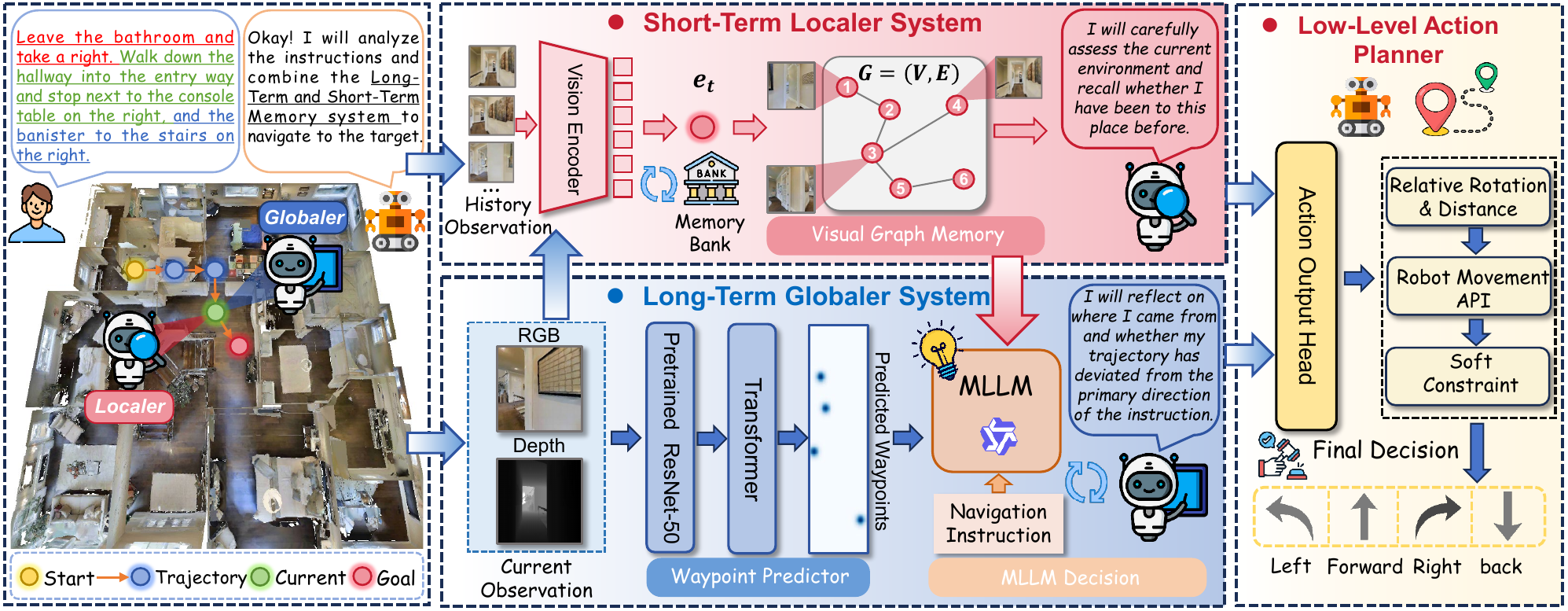}    \caption{\textbf{Workflow of the hierarchical memory system.} The visually driven Short-Term Localer mimics human spatial reasoning to detect revisits and reduce redundant exploration. The semantically driven Long-Term Globaler mirrors human global reflection to preserve origin awareness and directional alignment, ensuring long-horizon consistency.}
    \vspace{-3mm}
    \label{pipline} 
\end{figure*}

\subsection{Short-Term Localer System}
\label{Short-Term Localer System}

As discussed in Sec.~\ref{intro}, short-term Amnesia will lead to state drift and local looping issues. To address this challenge, we introduce the visually dominant Short-Term Localer system, as shown in Fig.~\ref{pipline}, which localizes the current observation at each step and provides executable exploration constraints to the MLLM, thereby stabilizing short-term state estimation and suppressing redundant exploration.

\textbf{Visual Graph Memory.} We maintain a visually grounded graph memory that evolves online during navigation, denoted as \( G_t = (V_t, E_t) \), where each node \( i \in V_t \) stores a triplet \( (\mathbf{v}_i, \tau_i, c_i) \). Here, \( \mathbf{v}_i \in \mathbb{R}^{512} \) represents the appearance embedding of the location, \( \tau_i \) denotes the most recent visit step, and \( c_i \) indicates the visit count. The edges \( E_t \) capture the connectivity between consecutive time steps. For the current multi-view RGB observation set \( \{I_t^d\} \), we employ a frozen CLIP image encoder \( \phi(\cdot) \) \cite{sigclip} to extract visual features and store them in a memory bank. To enhance robustness, Localer uses a forward-biased multi-view aggregation strategy: selecting directions from \( \mathcal{D}_k \), aggregating the embeddings of each view by weighted summation, and normalizing to obtain the current location embedding \( \mathbf{e}_t \), formulated as:
\begin{equation}
\mathbf{e}_t = \text{Norm}\left( \sum_{d \in \mathcal{D}_k} w_d \cdot \phi(I_t^d) \right),
\end{equation}
the forward-facing view is given higher weight to highlight regions related to the agent's movement.

\textbf{Localization and Graph Update.} At each step, Localer performs cosine similarity matching between the current observation embedding \( \mathbf{e}_t \) and the historical node embeddings stored in the memory bank:
\begin{equation}
s_i = \cos(\mathbf{e}_t, \mathbf{v}_i), \quad i \in V_t, \quad i^* = \arg\max_i s_i,
\end{equation}
When the matching similarity \( s_{i^*} \geq \theta_t \)(initialized to 0.85), the current position is considered to revisit node \( i^* \), completing the localization process. Otherwise, a new node is created. To balance the trade-off between early map building and later revisit detection, we employ a piecewise adaptive threshold \( \theta_t \): If the memory nodes are sparse, a higher threshold is used to strictly avoid merging different locations, and the threshold is gradually relaxed as the graph size increases to facilitate revisit detection. When a node is revisited, we apply a momentum update to stabilize its representation:
\begin{equation}
\mathbf{v}_{i^*} \leftarrow \text{Norm}\left( (1 - \alpha) \mathbf{v}_{i^*} + \alpha \mathbf{e}_t \right), \quad \alpha = 0.15,
\end{equation}
Simultaneously, we update the visit step \( \tau_{i^*} \leftarrow t \) and the visit count \( c_{i^*} \leftarrow c_{i^*} + 1 \). Finally, an undirected edge is added between adjacent nodes \( (u_{t-1}, u_t) \), resulting in the updated graph structure \( G_t \). This process enables each step to provide a discrete short-term state identifier \( u_t \), thereby facilitating state calibration through visual retrospection.

\textbf{Short-term memory injection.} During the decision-making phase, the Localer computes the appearance embedding \( \mathbf{e}_t^a = \phi(I_t^a) \) for each direction in the candidate direction set \( \mathcal{A}_t \), and estimates the degree of revisit of each direction by calculating the maximum similarity with the current graph memory \( V_t \), thereby obtaining the novelty of each direction \( n_t(a) \in [0, 1] \). A higher similarity indicates lower novelty. In the action space, nodes with higher visit counts and lower novelty are filtered or deprioritized, and the candidate direction priorities are output using a soft constraint strategy, which serves as exploration guidance. Localer transforms this short-term state information into a textual context that can be directly utilized by the MLLM and appends it to the multimodal navigation prompts. The format is summarized as follows:
\begin{leftbar}
\noindent
{\small \textit{\textbf{User:}} \textit{\textbf{Please summarize the short-term memory, including the number of explored locations, revisit statistics, and priorities of the candidate directions, while prioritizing unexplored directions without violating the instruction.}}}
\end{leftbar}

By explicitly providing visual localization signals and exploration soft constraints to the MLLM, the Short-Term Localer significantly mitigates the issue of short-term amnesia without relying on additional supervision, reducing local loops and redundant exploration, thus making subsequent decisions more stable and reliable.

\subsection{Long-Term Globaler System}
\label{Long-Term Globaler System}
In addition to short-term memory focusing on sequential decisions in the immediate neighborhood, navigation requires long-term memory to maintain stable representations of historical trajectories and support global planning. To address this, we propose the semantically driven Long-Term Globaler system, as illustrated in Fig.~\ref{pipline}, which recalls global context to calibrate local actions, thereby improving long-term consistency.

\textbf{Global Extraction and Origin Tracking.} At the beginning of an episode, Globaler extracts the global navigation schema from the instruction using a single MLLM textual inference, denoted as:
\begin{equation}
S = \{\text{PrimaryDir}, \text{FinalTarget}, \text{NavPattern}\},
\end{equation}
where \(\text{PrimaryDir}\) represents the general directional bias, \(\text{FinalTarget}\) is the target landmark, and \(\text{NavPattern}\) briefly describes the overall movement strategy. By structuring the extraction of the instruction goals, a stable global anchor is obtained. This global schema \(S\) is then cached as long-term semantic memory, guiding the entire decision-making process. Furthermore, to strengthen the agent's global orientation awareness, Globaler explicitly maintains the agent’s \(\text{CameFrom}\) variable, which indicates the relative return direction based on the current orientation. Let the actual movement direction at step \(t\) be \(a_t\), then the return direction for the next step is updated as \(\text{CameFrom}_{t+1} = \text{Opp}(a_t)\), where \(\text{Opp}(\cdot)\) denotes the reverse direction mapping. Finally, Globaler records a summary of the last \(k\) steps (\(k = 5\)), forming a compact global state overview: \(\{a_{t-k}, \dots, a_t\}\).

\textbf{Long-term memory injection.} At each decision-making step, Globaler compresses long-term memory into context information directly usable by the MLLM, which is then appended to the multimodal navigation prompts:
\begin{leftbar}
\noindent
{\small \textit{\textbf{User:}} \textit{\textbf{Please review the direction you came from \textless \textbf{CameFrom}\textgreater. and check if the current direction aligns with the global state, including \textless \textbf{PrimaryDir}\textgreater, \textless \textbf{FinalTarget}\textgreater, and \textless \textbf{NavPattern}\textgreater.}}}
\end{leftbar}

This process complements the Short-term Localer’s visual retrospection by providing stable target anchoring and directional bias, ensuring that local actions remain aligned with global intent without expanding the context window. It further mitigates drift from accumulated long-range noise, thereby enhancing overall navigation stability.

\section{Experiments}
\label{Experiments}

\begin{figure}[!t] 
    \centering 
    \includegraphics[width=\columnwidth]{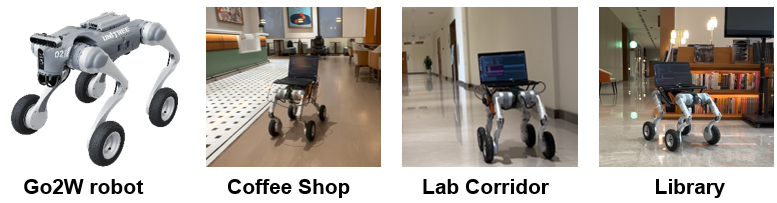}    
    \caption{The Go2W robot and the arrangement of the real-world environment.}
    \vspace{-4mm}
    \label{real-envir} 
\end{figure}

\begin{figure*}[!t] 
    \centering 
    \includegraphics[width=\textwidth]{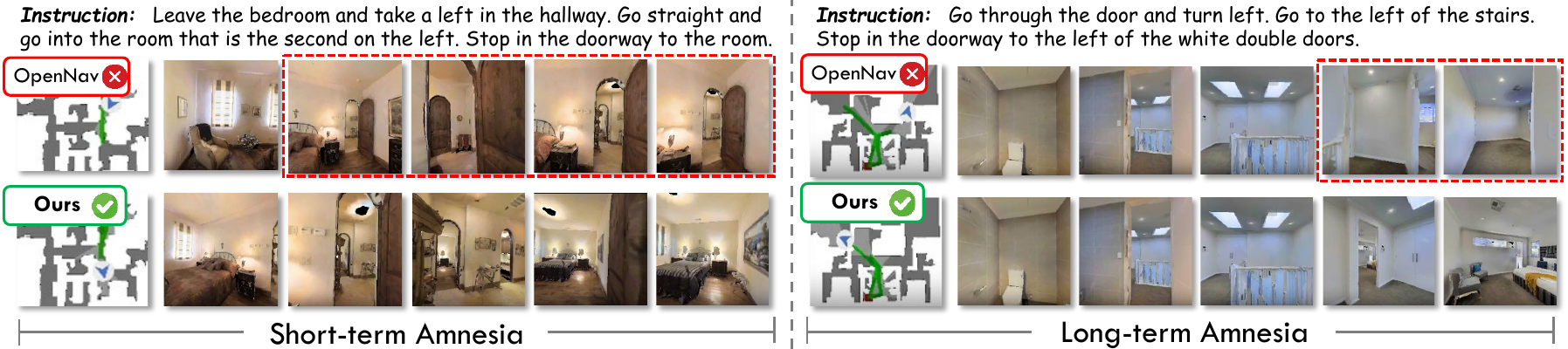}    \caption{\textbf{Qualitative results of OpenNav and HiMemVLN in simulation environments.} In contrast to the state-of-the-art open-source method , OpenNav, which exhibits short-term and long-term amnesia reflected by looping and deviating behaviors in the \textcolor{red}{red} dashed boxes, our method eliminates navigation amnesia and accurately follows the given instructions.}
    \vspace{-3mm}
    \label{expfig1} 
\end{figure*}

\subsection{Experiment Setup}
\noindent \textbf{Simulated Environments.} Our implementation is built upon the Habitat simulator \cite{habitat}, and the waypoint predictor is evaluated on the MP3D val-unseen split following the standardized evaluation protocol \cite{navid}. In addition, we benchmark the navigator over 100 episodes under the evaluation frameworks of Open-Nav \cite{opennav} and SmartWay \cite{smartway}, comparing against both learning-based VLN methods and LLM or MLLM approaches to ensure direct and fair comparability.

\noindent \textbf{Real-world Environments.} To further evaluate our approach in real-world settings, we design comprehensive experiments across indoor environments with varying spatial complexity, as shown in Fig.~\ref{real-envir}. We annotate 20 navigation instructions, including both simple and complex cases, to assess the agent’s performance on basic and advanced navigation scenarios. All real-world evaluations are conducted on a workstation equipped with dual NVIDIA L40 GPUs.

\noindent \textbf{Implementation Details.} In simulated experiments, we adopt the waypoint predictor pretrained in [18]. 
For real-world experiments, we deploy our system on a Unitree Go2W EDU wheeled-legged robot, which has 16 degrees of freedom (DoFs), including 12 actuated leg joints and 4 independently driven wheels, as shown in Fig.~\ref{real-envir}.
Our deployment follows a distributed two-stage architecture. 
The onboard Intel RealSense D435i continuously captures RGB-D observations, which are streamed to a local workstation equipped with an NVIDIA RTX 4090 GPU for preprocessing and data relay. The streamed visual observations are fed into an LLM-based navigator for high-level decision making.
For the LLM navigator, we use vLLM to deploy and test four open-source multimodal large models: Qwen2-72B-Instruct \cite{qwen2}, LLaVA-OV-1.5-8B \cite{llava}, InternVL-3.5-8B \cite{internvl3}, and Qwen2.5-VL-7B \cite{qwen25}. 
For practical deployment efficiency, Qwen2-72B-Instruct is implemented using 4-bit quantization. 
These models are executed on a remote server equipped with dual NVIDIA L40 GPUs, allowing efficient inference while maintaining real-time responsiveness.

\noindent \textbf{Evaluation Metrics.} Following prior work \cite{canav}, we evaluate navigation performance using standard VLN metrics, including success rate (SR), oracle success rate (OSR), normalized Dynamic Time Warping (nDTW), success weighted by path length (SPL), trajectory length (TL), and navigation error (NE). Consistent with the evaluation protocols in \cite{opennav,smartway}, an episode is considered successful if the agent stops within 3 meters of the goal in VLN-CE and within 2 meters in real-world settings. These criteria ensure balanced and fair evaluation across diverse environments while accounting for both navigational precision and spatial constraints.

\begin{table}[!t]
\centering
\caption{\textbf{Comparison on simulated environment R2R-CE.} Relevant reported results are sourced from \cite{smartway,opennav}.}
\renewcommand{\arraystretch}{1.1}
\setlength{\tabcolsep}{1.5pt}
\resizebox{\columnwidth}{!}{
\begin{tabular}{c|cccccc}
\hline
\textbf{Method} & \textbf{TL} & \textbf{NE$\downarrow$} & \textbf{nDTW$\uparrow$} & \textbf{OSR$\uparrow$} & \textbf{SR$\uparrow$} & \textbf{SPL$\uparrow$} \\
\hline

\rowcolor[HTML]{E5E5E5}
\multicolumn{7}{c}{\textbf{Supervised Learning}} \\
\hline
\rowcolor[HTML]{E5E5E5}
CMA \cite{cma}          & 11.08 & 6.92 & 50.77 & 45 & 37 & 32.17 \\
\rowcolor[HTML]{E5E5E5}
RecBERT \cite{bevbert}   & 11.06 & 5.8  & 54.81 & 57 & 48 & 43.22 \\
\rowcolor[HTML]{E5E5E5}
BEVBert \cite{bevbert} & 13.63 & 5.13 & 61.40 & 64 & 60 & 53.41 \\
\rowcolor[HTML]{E5E5E5}
ETPNav \cite{etpnav} & 11.08 & 5.15 & 61.15 & 58 & 52 & 52.18 \\
\hline

\multicolumn{7}{c}{\textbf{Close-Source Model Zero-shot Navigation}} \\
\hline
LXMERT \cite{lxmert}    & 15.79 & 10.48 & 18.73 & 22.00 & 2  & 1.87  \\
DiscussNav-GPT4 \cite{opennav} & 6.27  & 7.77  & 42.87 & 15 & 11 & 10.51 \\
MapGPT-CE-GPT4o \cite{smartway} & 12.63  & 8.16  & --- & 21 & 7 & 5.04 \\
Open-Nav-GPT4 \cite{opennav}      & 7.68  & 6.70 & 45.79 & 23 & 19 & 16.10 \\
SmartWay \cite{smartway}   & 13.09  & 7.01 & --- & 51  & 29 & 22.46 \\
\hline

\multicolumn{7}{c}{\textbf{Open-Source Model Zero-shot Navigation}} \\
\hline
Open-Nav-Llama3.1   & 8.07  & 7.25 & 44.99 & 23 & 16 & 12.90 \\
Open-Nav-Qwen2-72B   & 7.21  & 8.14 & 43.14 & 23 & 14 & 12.11 \\
\rowcolor{lightblue} \textbf{HiMemVLN-Qwen2-VL-72B}       & 7.55 & \textbf{6.65} & \textbf{52.79} & \textbf{36} & \textbf{30} & \textbf{26.85} \\
$\Delta$& {\color{red}-0.34}  & {\color{mygreen}-1.49}& {\color{mygreen}+9.65} &{\color{mygreen}+13} &{\color{mygreen}+16} & {\color{mygreen}+14.74} \\
\hline
\end{tabular}
}
\vspace{-6mm}
\label{tab1}
\end{table}

\begin{figure}[!t] 
    \centering 
    \includegraphics[width=0.9\columnwidth]{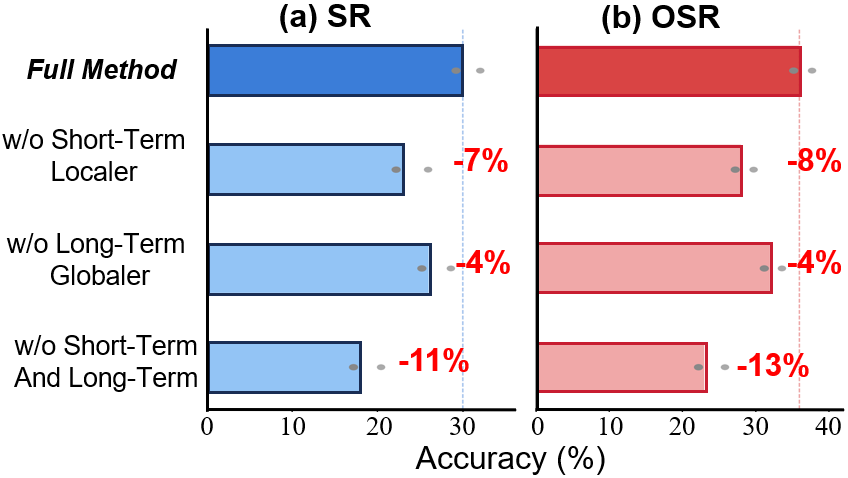}    
    \caption{Ablation study of different components.} 
    \label{figexp3}
    \vspace{-3mm}
\end{figure}

\subsection{Comparison on Simulated Environment}
\label{Comparison on Simulated Environment}
We systematically compare HiMemVLN with mainstream VLN approaches. To ensure fairness, all methods employ the same waypoint predictor, focusing on the performance of the navigator itself. As shown in Table~\ref{tab1}, methods are categorized into supervised, closed-source zero-shot, and open-source zero-shot groups. Zero-shot methods, which do not undergo task-specific fine-tuning, typically exhibit lower SR and SPL compared to supervised learning approaches. Among them, closed-source methods benefit from stronger reasoning and contextual modeling capabilities, resulting in relatively better performance than open-source models. However, they rely on costly API calls and pose security risks. Under a strict open-source zero-shot setting, HiMemVLN significantly outperforms OpenNav, attaining an OSR of 36\%, an SR of 30\%, and an SPL of 26.85\%. It not only surpasses open-source baselines by a large margin but also approaches or exceeds several closed-source models. Additionally, we visualize several navigation processes in the simulation environment in Fig.~\ref{expfig1}. The SOTA open-source method, OpenNav \cite{opennav}, exhibits navigation amnesia, causing the agent to circle in place and deviate from the main intent of the instruction. In contrast, our method demonstrates a strong understanding of the task, effectively perceiving and executing each sub-instruction until reaching the correct target location. These results highlight the effectiveness of the hierarchical memory mechanism in enhancing environmental understanding and decision-making, substantially bridging the gap between open-source zero-shot methods and supervised or closed-source approaches.

\subsection{Comparison on Real-world Environments}
\label{Comparison on Real-world Environments}
\begin{figure*}[h]
\centering
\includegraphics
[width=1\textwidth]
{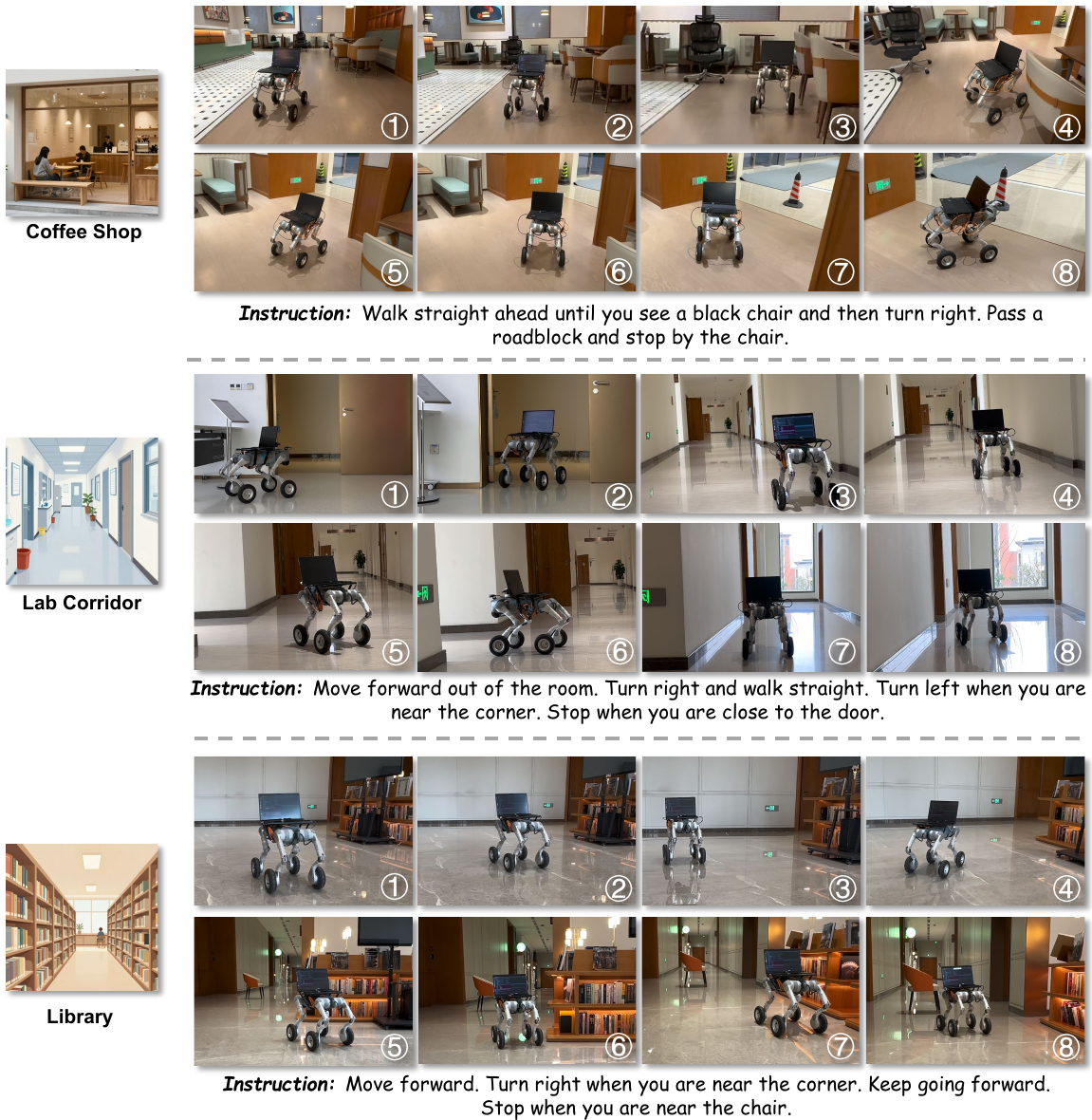} 
\caption{
\textbf{Real-world deployment of HiMemVLN in three representative indoor environments.} The robot successfully executes long-horizon task sequences and demonstrates robust navigation in complex settings.
}
\label{fig:real_exp}
\end{figure*}

We conduct further experiments in real-world environments, and Fig.~\ref{fig:real_exp} presents several visualizations of our method. HiMemVLN outperforms OpenNav in success rate, with OpenNav achieving SR and NE of 18 and 4.27, respectively, while our method achieves 32 (↑14) and 3.54 (↓0.73).  This improvement stems from the need for fine-grained scene perceptionand directional localization capabilities in challenging real-world environments. HiMemVLN’s hierarchical memory system equips it with these capabilities, resulting in improved generalization.

\begin{table}[!t]
\centering
\renewcommand{\arraystretch}{1.1}
\caption{Performance of Different Open-Source LLMs on Navigation.}
\resizebox{\columnwidth}{!}{
\begin{tabular}{c|cccccc}
\hline
\textbf{Method} & \textbf{TL} & \textbf{NE$\downarrow$} & \textbf{nDTW$\uparrow$} & \textbf{OSR$\uparrow$} & \textbf{SR$\uparrow$} & \textbf{SPL$\uparrow$} \\ \hline
Qwen2-VL-72B    & 7.55        & 6.65                    & 52.79                   & 36                    & 30                   & 26.85                  \\ 
InternVL-3.5-8B & 7.75        & 7.41                    & 49.91                   & 28                    & 23                   & 18.29                  \\ 
LLaVA-OV-1.5-8B & 7.36        & 7.58                    & 47.39                   & 26                    & 20                   & 17.72                  \\ 
Qwen2.5-VL-7B   & 7.32        & 7.67                    & 47.04                   & 23                    & 15                   & 12.16                  \\ 
\hline
\end{tabular}
}
\label{tab2}
\end{table}

\subsection{Ablation Study}
\label{Ablation Study}

\noindent \textbf{The impact of different components.} We conducted ablation studies to evaluate the impact of different components on navigation (Fig.~\ref{figexp3}). Without the multimodal LLM, the agent’s environmental perception significantly degraded, resulting in poor performance. Removing the Short-Term Localer System caused a decline, with the agent becoming trapped in local loops due to the lack of short-term visual feedback. Similarly, removing the Long-Term Globaler System led to a loss of orientation in long-term planning, causing navigation failure and reduced accuracy. When both memory systems are combined, the agent shows improved environmental awareness and decision-making, achieving optimal performance.

\noindent \textbf{Comparison of Different MLLMs on Navigation.} 
To evaluate the decision-making ability of other open-source MLLMs as navigators, we assessed the navigation performance of four MLLMs, as shown in Table~\ref{tab2}. Among them, Qwen2-VL-72B performed the best, followed by InternVL-3.5-8B. This comparison provides benchmark results and aims to inspire further research and exploration in the field of zero-shot navigation with open-source MLLMs.

\section{Conclusions}
\label{Conclusions}

This paper proposes HiMemVLN, a framework designed to enhance the reliability of open-source zero-shot VLN. It addresses the navigation amnesia issue by introducing a hierarchical memory system that stabilizes state estimation and long-horizon decision making. Inspired by principles of human cognition, HiMemVLN integrates a visually dominant short-term Localer and a semantically driven long-term Globaler. The Localer leverages online visual graph memory for state localization and revisit suppression, while the Globaler enforces global instruction consistency to prevent long-horizon decision drift. Extensive experiments in both simulated and real-world settings demonstrate that HiMemVLN achieves SOTA performance and substantially narrows the gap to closed-source models, validating its practicality under privacy-preserving and deployment-oriented constraints.

\bibliographystyle{IEEEtran} 
\bibliography{references}  

@inproceedings{vln1,
  title={Vision-and-language navigation: Interpreting visually-grounded navigation instructions in real environments},
  author={Anderson, Peter and Wu, Qi and Teney, Damien and Bruce, Jake and Johnson, Mark and S{\"u}nderhauf, Niko and Reid, Ian and Gould, Stephen and Van Den Hengel, Anton},
  booktitle={Proceedings of the IEEE conference on computer vision and pattern recognition},
  pages={3674--3683},
  year={2018}
}

@inproceedings{vlnce,
  title={Beyond the nav-graph: Vision-and-language navigation in continuous environments},
  author={Krantz, Jacob and Wijmans, Erik and Majumdar, Arjun and Batra, Dhruv and Lee, Stefan},
  booktitle={European Conference on Computer Vision},
  pages={104--120},
  year={2020},
  organization={Springer}
}

@article{llm2,
  title={Gpt-4 technical report},
  author={Achiam, Josh and Adler, Steven and Agarwal, Sandhini and Ahmad, Lama and Akkaya, Ilge and Aleman, Florencia Leoni and Almeida, Diogo and Altenschmidt, Janko and Altman, Sam and Anadkat, Shyamal and others},
  journal={arXiv preprint arXiv:2303.08774},
  year={2023}
}

@article{gpt4o,
  title={Gpt-4o system card},
  author={Hurst, Aaron and Lerer, Adam and Goucher, Adam P and Perelman, Adam and Ramesh, Aditya and Clark, Aidan and Ostrow, AJ and Welihinda, Akila and Hayes, Alan and Radford, Alec and others},
  journal={arXiv preprint arXiv:2410.21276},
  year={2024}
}

@inproceedings{cma,
  title={Bridging the gap between learning in discrete and continuous environments for vision-and-language navigation},
  author={Hong, Yicong and Wang, Zun and Wu, Qi and Gould, Stephen},
  booktitle={Proceedings of the IEEE/CVF conference on computer vision and pattern recognition},
  pages={15439--15449},
  year={2022}
}

@article{bevbert,
  title={Bevbert: Multimodal map pre-training for language-guided navigation},
  author={An, Dong and Qi, Yuankai and Li, Yangguang and Huang, Yan and Wang, Liang and Tan, Tieniu and Shao, Jing},
  journal={arXiv preprint arXiv:2212.04385},
  year={2022}
}

@article{etpnav,
  title={Etpnav: Evolving topological planning for vision-language navigation in continuous environments},
  author={An, Dong and Wang, Hanqing and Wang, Wenguan and Wang, Zun and Huang, Yan and He, Keji and Wang, Liang},
  journal={IEEE Transactions on Pattern Analysis and Machine Intelligence},
  year={2024},
  publisher={IEEE}
}

@article{navid,
  title={Navid: Video-based vlm plans the next step for vision-and-language navigation},
  author={Zhang, Jiazhao and Wang, Kunyu and Xu, Rongtao and Zhou, Gengze and Hong, Yicong and Fang, Xiaomeng and Wu, Qi and Zhang, Zhizheng and Wang, He},
  journal={arXiv preprint arXiv:2402.15852},
  year={2024}
}

@article{canav,
  title={Constraint-aware zero-shot vision-language navigation in continuous environments},
  author={Chen, Kehan and An, Dong and Huang, Yan and Xu, Rongtao and Su, Yifei and Ling, Yonggen and Reid, Ian and Wang, Liang},
  journal={IEEE Transactions on Pattern Analysis and Machine Intelligence},
  year={2025},
  publisher={IEEE}
}

@inproceedings{navgpt,
  title={Navgpt: Explicit reasoning in vision-and-language navigation with large language models},
  author={Zhou, Gengze and Hong, Yicong and Wu, Qi},
  booktitle={Proceedings of the AAAI Conference on Artificial Intelligence},
  volume={38},
  number={7},
  pages={7641--7649},
  year={2024}
}

@inproceedings{discussnav,
  title={Discuss before moving: Visual language navigation via multi-expert discussions},
  author={Long, Yuxing and Li, Xiaoqi and Cai, Wenzhe and Dong, Hao},
  booktitle={2024 IEEE International Conference on Robotics and Automation (ICRA)},
  pages={17380--17387},
  year={2024},
  organization={IEEE}
}

@article{llama,
  title={Lawyer llama technical report},
  author={Huang, Quzhe and Tao, Mingxu and Zhang, Chen and An, Zhenwei and Jiang, Cong and Chen, Zhibin and Wu, Zirui and Feng, Yansong},
  journal={arXiv preprint arXiv:2305.15062},
  year={2023}
}

@inproceedings{opennav,
  title={Open-nav: Exploring zero-shot vision-and-language navigation in continuous environment with open-source llms},
  author={Qiao, Yanyuan and Lyu, Wenqi and Wang, Hui and Wang, Zixu and Li, Zerui and Zhang, Yuan and Tan, Mingkui and Wu, Qi},
  booktitle={2025 IEEE International Conference on Robotics and Automation (ICRA)},
  pages={6710--6717},
  year={2025},
  organization={IEEE}
}

@inproceedings{smartway,
  title={Smartway: Enhanced waypoint prediction and backtracking for zero-shot vision-and-language navigation},
  author={Shi, Xiangyu and Li, Zerui and Lyu, Wenqi and Xia, Jiatong and Dayoub, Feras and Qiao, Yanyuan and Wu, Qi},
  booktitle={2025 IEEE/RSJ International Conference on Intelligent Robots and Systems (IROS)},
  pages={16923--16930},
  year={2025},
  organization={IEEE}
}

@inproceedings{lxmert,
  title={Lxmert: Learning cross-modality encoder representations from transformers},
  author={Tan, Hao and Bansal, Mohit},
  booktitle={Proceedings of the 2019 conference on empirical methods in natural language processing and the 9th international joint conference on natural language processing (EMNLP-IJCNLP)},
  pages={5100--5111},
  year={2019}
}

@article{openclose1,
  title={Deepseek llm: Scaling open-source language models with longtermism},
  author={Bi, Xiao and Chen, Deli and Chen, Guanting and Chen, Shanhuang and Dai, Damai and Deng, Chengqi and Ding, Honghui and Dong, Kai and Du, Qiushi and Fu, Zhe and others},
  journal={arXiv preprint arXiv:2401.02954},
  year={2024}
}

@inproceedings{openclose2,
  title={A literature survey on open source large language models},
  author={Kukreja, Sanjay and Kumar, Tarun and Purohit, Amit and Dasgupta, Abhijit and Guha, Debashis},
  booktitle={Proceedings of the 2024 7th International Conference on Computers in Management and Business},
  pages={133--143},
  year={2024}
}

@article{theory1,
  title={The hippocampal memory indexing theory.},
  author={Teyler, Timothy J and DiScenna, Pascal},
  journal={Behavioral neuroscience},
  volume={100},
  number={2},
  pages={147},
  year={1986},
  publisher={American Psychological Association}
}

@article{theory2,
  title={Short-term memory and long-term memory are still different.},
  author={Norris, Dennis},
  journal={Psychological bulletin},
  volume={143},
  number={9},
  pages={992},
  year={2017},
  publisher={American Psychological Association}
}

@article{mp3d,
  title={Matterport3d: Learning from rgb-d data in indoor environments},
  author={Chang, Angel and Dai, Angela and Funkhouser, Thomas and Halber, Maciej and Niessner, Matthias and Savva, Manolis and Song, Shuran and Zeng, Andy and Zhang, Yinda},
  journal={arXiv preprint arXiv:1709.06158},
  year={2017}
}

@inproceedings{lstm1,
  title={Neighbor-view enhanced model for vision and language navigation},
  author={An, Dong and Qi, Yuankai and Huang, Yan and Wu, Qi and Wang, Liang and Tan, Tieniu},
  booktitle={Proceedings of the 29th ACM International Conference on Multimedia},
  pages={5101--5109},
  year={2021}
}

@article{lstm2,
  title={Language and visual entity relationship graph for agent navigation},
  author={Hong, Yicong and Rodriguez, Cristian and Qi, Yuankai and Wu, Qi and Gould, Stephen},
  journal={Advances in Neural Information Processing Systems},
  volume={33},
  pages={7685--7696},
  year={2020}
}

@inproceedings{rep1,
  title={Towards learning a generic agent for vision-and-language navigation via pre-training},
  author={Hao, Weituo and Li, Chunyuan and Li, Xiujun and Carin, Lawrence and Gao, Jianfeng},
  booktitle={Proceedings of the IEEE/CVF conference on computer vision and pattern recognition},
  pages={13137--13146},
  year={2020}
}

@inproceedings{rep2,
  title={Hop: History-and-order aware pre-training for vision-and-language navigation},
  author={Qiao, Yanyuan and Qi, Yuankai and Hong, Yicong and Yu, Zheng and Wang, Peng and Wu, Qi},
  booktitle={Proceedings of the IEEE/CVF Conference on Computer Vision and Pattern Recognition},
  pages={15418--15427},
  year={2022}
}

@inproceedings{regnav,
  title={Regnav: Room expert guided image-goal navigation},
  author={Li, Pengna and Wu, Kangyi and Fu, Jingwen and Zhou, Sanping},
  booktitle={Proceedings of the AAAI Conference on Artificial Intelligence},
  volume={39},
  number={5},
  pages={4860--4868},
  year={2025}
}

@article{instructnav,
  title={Instructnav: Zero-shot system for generic instruction navigation in unexplored environment},
  author={Long, Yuxing and Cai, Wenzhe and Wang, Hongcheng and Zhan, Guanqi and Dong, Hao},
  journal={arXiv preprint arXiv:2406.04882},
  year={2024}
}

@inproceedings{llmplanner,
  title={Llm-planner: Few-shot grounded planning for embodied agents with large language models},
  author={Song, Chan Hee and Wu, Jiaman and Washington, Clayton and Sadler, Brian M and Chao, Wei-Lun and Su, Yu},
  booktitle={Proceedings of the IEEE/CVF international conference on computer vision},
  pages={2998--3009},
  year={2023}
}

@inproceedings{spatialbot,
  title={Spatialbot: Precise spatial understanding with vision language models},
  author={Cai, Wenxiao and Ponomarenko, Iaroslav and Yuan, Jianhao and Li, Xiaoqi and Yang, Wankou and Dong, Hao and Zhao, Bo},
  booktitle={2025 IEEE International Conference on Robotics and Automation (ICRA)},
  pages={9490--9498},
  year={2025},
  organization={IEEE}
}

@inproceedings{ram,
  title={Recognize anything: A strong image tagging model},
  author={Zhang, Youcai and Huang, Xinyu and Ma, Jinyu and Li, Zhaoyang and Luo, Zhaochuan and Xie, Yanchun and Qin, Yuzhuo and Luo, Tong and Li, Yaqian and Liu, Shilong and others},
  booktitle={Proceedings of the IEEE/CVF Conference on Computer Vision and Pattern Recognition},
  pages={1724--1732},
  year={2024}
}

@inproceedings{resnet,
  title={Deep residual learning for image recognition},
  author={He, Kaiming and Zhang, Xiangyu and Ren, Shaoqing and Sun, Jian},
  booktitle={Proceedings of the IEEE conference on computer vision and pattern recognition},
  pages={770--778},
  year={2016}
}

@article{depth,
  title={Dd-ppo: Learning near-perfect pointgoal navigators from 2.5 billion frames},
  author={Wijmans, Erik and Kadian, Abhishek and Morcos, Ari and Lee, Stefan and Essa, Irfan and Parikh, Devi and Savva, Manolis and Batra, Dhruv},
  journal={arXiv preprint arXiv:1911.00357},
  year={2019}
}

@inproceedings{sigclip,
  title={Sigmoid loss for language image pre-training},
  author={Zhai, Xiaohua and Mustafa, Basil and Kolesnikov, Alexander and Beyer, Lucas},
  booktitle={Proceedings of the IEEE/CVF international conference on computer vision},
  pages={11975--11986},
  year={2023}
}

@inproceedings{habitat,
  title={Habitat: A platform for embodied ai research},
  author={Savva, Manolis and Kadian, Abhishek and Maksymets, Oleksandr and Zhao, Yili and Wijmans, Erik and Jain, Bhavana and Straub, Julian and Liu, Jia and Koltun, Vladlen and Malik, Jitendra and others},
  booktitle={Proceedings of the IEEE/CVF international conference on computer vision},
  pages={9339--9347},
  year={2019}
}

@article{qwen2,
  title={Qwen2-vl: Enhancing vision-language model's perception of the world at any resolution},
  author={Wang, Peng and Bai, Shuai and Tan, Sinan and Wang, Shijie and Fan, Zhihao and Bai, Jinze and Chen, Keqin and Liu, Xuejing and Wang, Jialin and Ge, Wenbin and others},
  journal={arXiv preprint arXiv:2409.12191},
  year={2024}
}

@article{internvl3,
  title={Internvl3. 5: Advancing open-source multimodal models in versatility, reasoning, and efficiency},
  author={Wang, Weiyun and Gao, Zhangwei and Gu, Lixin and Pu, Hengjun and Cui, Long and Wei, Xingguang and Liu, Zhaoyang and Jing, Linglin and Ye, Shenglong and Shao, Jie and others},
  journal={arXiv preprint arXiv:2508.18265},
  year={2025}
}

@article{qwen25,
  title={Fine-Tuning the Qwen2. 5-VL Model for Intelligent Applications in the Electrical Domain},
  author={Yao, Song and Lv, Chunli and Zhu, Kun and Qiu, Xiaobin},
  journal={EAI Endorsed Transactions on Energy Web},
  volume={12},
  year={2024},
  publisher={European Alliance for Innovation (EAI)}
}

@article{llava,
  title={Llava-onevision-1.5: Fully open framework for democratized multimodal training},
  author={An, Xiang and Xie, Yin and Yang, Kaicheng and Zhang, Wenkang and Zhao, Xiuwei and Cheng, Zheng and Wang, Yirui and Xu, Songcen and Chen, Changrui and Zhu, Didi and others},
  journal={arXiv preprint arXiv:2509.23661},
  year={2025}
}

\end{document}